# Generating Decision Structures and Causal Explanations For Decision Making

Spencer Star

Département d'informatique, Université Laval
Cité Universitaire, Québec, Canada G1K 7P4

BITNET: STAR@LAVALVM1.BITNET     ARPANET: STAR@B.GP.CS.CMU.EDU

*ABSTRACT--This paper examines two related problems that are central to developing an autonomous decision-making agent, such as a robot. Both problems require generating structured representations from a database of unstructured declarative knowledge that includes many facts and rules that are irrelevant in the problem context. The first problem is how to generate a well-structured decision problem from such a database. The second problem is how to generate, from the same database, a well-structured explanation of why some possible world occurred.*

*In this paper it is shown that the problem of generating the appropriate decision structure or explanation is intractable without introducing further constraints on the knowledge in the database. The paper proposes that the problem search space can be constrained by adding knowledge to the database about causal relations between events. In order to determine the causal knowledge that would be most useful, causal theories for deterministic and indeterministic universes are proposed. A program that uses some of these causal constraints has been used to generate explanations about faulty plans. The program shows the expected increase in efficiency as the causal constraints are introduced.*

## 1. INTRODUCTION

Decision analysis starts with a description of a problem, translates that description into a structural representation, and then uses algorithms to evaluate the structure in order to find the problem's optimal solution. The algorithms are tedious but are well-enough understood so that they can be written as a program that will do the evaluation automatically. [Shacter 1986; Pearl 1986; Lauritzen and Spiegelhalter 1987; Henrion 1986]

The key missing element of the decision analysis technology is a set of rules and procedures that specify how to translate a problem description into a structural representation [Breese 1987]. Currently, the task of creating a structural representation from a declarative description that contains many facts and rules that are irrelevant in the problem context can be described as the art of decision analysis.

The problem of generating a structural representation for a decision maker can be shown to be closely related to the problem of generating an explanation of about why some possible world occurred. These two related problems should be central concerns for anyone wanting to developing an autonomous decision-making agent, such as a robot. Both problems require generating structured representations from a database of unstructured declarative knowledge that includes many facts and rules that are irrelevant in the problem context. When the structured representation is used to find the decision acts that will maximize a decision-making agent's utility, I call it a decision structure. When the agent generates a similar structure to explain ex post why a particular set of outcomes occurred I call the structure an explanation.

In this paper it is shown that at one point in time the number of possible outcomes that could be represented in a decision structure is exponentially complex in the number of events represented in the agent's database. Furthermore, when searching for the correct transition rules from one state to another over time, the number of possible rules is exponentially complex in the number of outcomes at each moment. The results show the problem to be intractable without introducing further constraints on the knowledge in the database. The paper then examines the related problem of inferring the best explanation of the causal factors that have led to the current state of the world. Inference from a known outcome to the best explanation is called abduction. The problem of abduction is shown to be of the same complexity as that of generating the appropriate decision structure. The paper proposes that the problem search space can be constrained by adding knowledge to the database about causal relations between events. In order to determine the causal knowledge that would be most useful, causal theories for deterministic and indeterministic universes are proposed. A program that uses some of these causal constraints has been used to generate explanations about faulty plans. The program shows the expected increase in efficiency as the causal constraints are introduced.

323

## 2. LEARNING A DECISION STRUCTURE

In this section I will first set out the elements of a well-structured dynamic decision problem for an autonomous decision-making agent such as a robot. By examining the difference between the structured representation of the decision problem and the unstructured database of knowledge in declarative form, we will determine what parts of the decision problem have to be learned by the agent.

The decision problem will then be recast in a possible-worlds framework and the complexity of the learning task will be described in terms of the size of the search space among the possible worlds. It will be shown that we must introduce a bias in terms of prior knowledge that constrains the search space. It is proposed that the constraints provided by encoding causal relations in the knowledge base will introduce sufficient bias to make the problem tractable.

*A structured representation of the decision problem.* A dynamic decision problem can be described in terms of a decision maker's search for an optimal plan. A *plan* consists of a goal, a sequence of decisions taken to achieve that goal, and the outcomes associated with those decisions. There are probabilities and utilities associated with each outcome, and the decision maker chooses decision acts that will maximize his expected utility. I will call a well-structured representation of this problem a *decision structure*. The decision problem of finding an optimal plan can be described as a dynamic maximization problem that has the following decision structure:

$$\text{Max } \Sigma_t \text{ EU}(D(t),X(t)) \text{ s.t. } X(t+1) = R(D(t), X(t))$$

where the variables and relevant features of the decision problem are defined as follows:

*(i) $D(t)$: Decision space at time $t$.* The finite space of potential acts. A decision maker can select a specific act $\{d(t)_i\}$ from the set of k possible decision acts $D(t) = \{d(t)_1, d(t)_2,..., d(t)_k\}$.

*(ii) $X(t)$: State space at time $t$.* The finite space of feasible outcomes of the world: $X(t) = \{x(t)_1, x(t)_2,..., x(t)_l\}$.

*(iii) $U(t)$: Utility evaluation in the state space $X(t)$.* For every decision $d(t)_i$ the decision maker assigns utility $U(d(t)_i,X(t))$ to each possible outcome in the state space at time t.

*(iv) $P(X(t)/d(t)_i)$: Probability assessment on $X(t)$ conditional on decision $d(t)_i$.* For every decision $d(t)_i$ the decision maker directly or indirectly assigns a joint probability measure $P(X(t)/d(t)_i)$ to each possible outcome in the state space at time t.

*(v) $R(D(t), X(t))$: Transition rules.* A set of rules that map each element of $X(t)$ and $D(t)$ into an element of $X(t+1)$.

*(vi) $EU(t)$: Expected utility at time $t$.* For each decision and all outcomes at time t the associated utilities are multiplied by the corresponding conditional probabilities. The sum over possible decisions and outcome at time t is expected utility at time t: $EU(t) = \Sigma_i U(d(t)_i,X(t))P(X(t)/d(t)_i)$

*Generating the Appropriate Decision Structure.* Now that we know what the elements of a decision structure are, let us see how an intelligent agent can go about generating such a structure. The basic approach that will be used is *generate and test*. The agent uses its knowledge base and machine time to generate a set of possible decision structures and then tests these structures to find the best one given its limited resources. We will provide the decision-making agent with the following *limited resource inputs*: K, an initial database of knowledge and beliefs; M, a computing machine; and O, observations of the outside world.

The agent uses these inputs to learn things, by reasoning and observing, that enable it to generate and identify the appropriate decision structure. Learning will be defined simply as any change to the knowledge base that permits an agent to achieve greater total utility, over some time interval, than could be obtained without learning. If a particular decision is to be made only one time, a learning program should be able to make some generalization from the problem structure that would be useful (utility increasing) when considering analogous decision problems.

324

For expository purposes, I will describe the kinds of things that are to be learned in terms of an influence diagram that graphically represents the decision structure. An agent can learn three classes of concepts:

*Learn the nodes.* Nodes, which I will also call *events*, consist of decision acts (the decision to act and the act itself) and outcomes. It is assumed that the agent's knowledge base can describe a large finite number of nodes or events, only a small subset of which will be relevant to a particular problem. The determination of the appropriate context and the relevant decisions and outcomes requires searching through the space of all possible events. In order to make this problem tractable, knowledge must be introduced that indicates those events that are most likely to be relevant in the particular context. In terms of an influence diagram, let there be a set of pre-determined nodes that can be used. Choosing the appropriate subset is learning the nodes.

*Learn the arcs.* Define this to mean learning the topology of the minimal number of arcs that must be drawn in an influence diagram together with the transition rules. The main role of arcs in Bayes nets and influence diagrams is to indicate by their presence and absence specific independencies and conditional independencies among the nodes. Arcs in influence diagrams correspond somewhat loosely to the transition rules of the decision structure. The standard transition rule used in influence diagrams is Bayes' rule for calculating inverse probabilities. However, transition rules in a formal decision structure contain explicit references to time while arcs are ambiguous with respect to time. Arcs into a decision node show that the information flowing along the arc is available before the decision is made. Arcs between outcome nodes, however, do not clearly assume that the prior node is prior in time. They can indicate functional and definitional relationships between nodes at one point in time or they can indicate, sometimes only implicitly, transitions between time periods. This ambiguity can lead to paradoxes such as Newcombe's problem [Nozick 1969; Gardenfors and Sahlin 1988].

It should be emphasized that up to this point we have been concerned only with learning *qualitative* and *symbolic* structures--the topology of a decision structure--which is perhaps the most difficult and least understood aspect of decision analysis.

*Learn the numbers.* I assume that the particular decision problem to be faced is not known with certainty before it occurs, thus making it impossible to enter into the database all the appropriate probabilities and utilities. However, there will be included in the database certain reference decision problems with associated values for probabilities and utilities. One way to learn the numbers is for the agent to use analogical reasoning, comparing the symbolic structure of the problem at hand, called the target problem, with the symbolic structure of the reference problems that it knows about. The agent will then make estimates of the missing values, basing its reasoning on the strength and type of associations that are discovered. The agent may also make use of default rules such as equal probabilities among events in the basic reference class.

Note that the "equal probabilities" assumption is strongly dependent on the symbolic structure that the agent has generated. The determination of the number of homogeneous groups in the basic reference class is part of the problem of learning the nodes. Obviously the probabilities that an agent will assign to equally probable events is an inverse function of the number of possible outcomes that are considered relevant.

*A possible-worlds approach.* In this section I recast the decision problem in terms of a set of possible worlds to provide the basis for an analysis of problem complexity. It is assumed that sentences in standard first order predicate logic are used to describe decision acts, outcomes, transition rules, and quantitative values. The logic may be augmented if necessary to represent probability theory [Nilsson 1986].

Throughout this paper I will be discussing events and the causal relations among events. Define the union of the set of decision acts and the set of outcomes as the set of *events*. Let there be a fixed set of m possible events that the agent may have to deal with. The event set is sufficient to describe any decision problem that the decision agent will face, but not all of the events are relevant for any particular decision problem. Let $S = \{s_1, s_2, ..., s_m\}$ be the set of sentences describing the m events at time t. The time-period index is suppressed except when necessary to distinguish between different time periods. Define a sentence to be true *with respect to a possible world* only when it evaluates to true using the interpretation associated with that possible world.

Define the set of possible worlds at time t as the power set of S. Each subset of S describes a possible world $w(t)_i$ at time t and every possible combination of events is represented in some possible world. Represent the set of possible worlds at t by the binary-valued matrix $W = [w(t)_{ij}]$, where i

325

represents the rows and j the columns. Let the jth column be the sentence corresponding the jth event, and the ith row be the ith possible world. Those events that are true in the ith possible world will be represented by ones in the appropriate columns in row i.

Generating all possible worlds and identifying the one closest to the agent's actual world is exponentially complex in the number of events. Given m possible events, the number of possible worlds at time t is $n = 2^m$. But this does not consider the dynamic problem of moving from one possible world at t to another at $t+1$. This is the problem of learning the transition rules, or, loosely speaking, learning the arcs. The problem can be analyzed as follows.

Let $d(t)_i$ be a decision act that the agent might decide to perform. Assume that the decision to act and the performance of the act occur simultaneously. Define a *decision-world* as a possible world which is to some degree like the actual world before t and in which the agent decides to do $d(t)_i$ at t. Let $dw(t)_i^*$ be the decision world which at t is the *nearest world to the real world*. The agent has chosen to do what he believes to be the utility maximizing decision act based on his prediction of the possible worlds (and their probabilities and utilities) that were accessible from the set of decision worlds.

Due to ignorance about the real world and limited computational resources, there will not be only one possible world at $t+1$, but many. Let the agent have generated the possible-worlds matrix $W(t+1)$. Define the *transition relation*, $r(dw(t)_i^*, W(t+1)_i)$, between the closest decision-world at time t and the ith subset of possible worlds $W(t+1)_i$ that $dw(t)_i^*$ could be transformed into. Represent the possible world closest to the actual world resulting from $dw(t)_i^*$ as $w(t+1)_i^*$.

If the universe was deterministic and the agent had perfect information and was omniscient, then it would know the transition relation $r(dw(t)_i^*, w(t+1)_i^*)$. Limited resources and imperfect information force the agent to be satisfied with a generalization of $w(t+1)_i^*$, represented by some subset of possible worlds. Obviously there is a tradeoff in decision making between using the most general representation of the future (all possible worlds) and using the most specific representation (only one possible world).

Choosing the proper degree of generalization means choosing how many possible outcomes will adequately describe the possible states of the world after making a decision. This is a different problem than learning the nodes. It is concerned with learning a computationally tractable transition relation: in essence, learning the transition rules.

The combinatorics of the agent's problem of learning the transition rules is exponential in the number of possible worlds. For example, suppose there is only one possible world that is accessible from $dw(t)_i^*$. There are m possible transition rules, one for each possible world at $t+1$. If there are two possible worlds accessible from $dw(t)_i^*$, each different combination of two worlds corresponds to a different transition rule. Since there are n possible worlds, there are $2^n$ possible transition relations. Recall that n itself was the result of $2^m$ possible combinations of events. Thus if, for example, our robot operates in a microworld of four potential chance events and one decision event at time t, there are five events and $32 = 2^5$ possible worlds. Suppose that the nearest decision world $dw(t)_i^*$ has been identified. From among the 32 possible worlds in $t+1$, there are $2^{32}$ different, not necessarily exclusive, transition rules. With this kind of combinatorial explosion, generate and test is clearly a computationally intractable method.

What is needed is a way of organizing the space of nodes and arcs so that only a small subset of possible worlds and transition relations need be considered. In machine learning, an effective technique for learning concepts has been to organize the search space hierarchically from the most general to the most specific concepts. Define the degree of generalization to be a function of the number of distinct possible worlds that would be included under the generalization. For n possible worlds, there are $2^n$ transition relations divided into n levels of generality. The result is that each level of generality has on the average $2^{n-1}$ transition relations, a shallow and wide hierarchy that does not help in dealing with the combinatorial explosion.

The problem of making sense of complexity is closely related to the problems of predicting events and explaining why events occur. One widely accepted view of prediction and explanation sees them as inverse functions: an event predicted with a high probability of occurring can be explained by factors yielding the prediction. An alternative view, which will be presented later, introduces causal relations. If the causal relations leading to a high-probability event are understood as well those leading to a low-probability event, the explanations have similar power. Thus, explanation is more than the identification of regularities that enable one to generalize from a sample population because it involves theoretical reasoning about causal processes that link events. Decision making is at its very foundations a based on the search for causal relations. Decisions set in motion causal processes that determine which of the possible worlds will occur.

A framework for explanation can be built very easily on the possible worlds foundation that we have developed. Consider the possible world, $w(t)_i^*$, which most closely resembles the world we are actually in. To put this in a problem context, suppose the decision agent at t-1 has made an optimal plan that predicted a different world as the most likely outcome at time t. Suppose further that the realized world was considered very unlikely and let realized utility be much lower than expected utility. Is the unexpected outcome due to a flaw in the plan or is it one of those occurrences of relatively rare outcomes that can be expected in a world of imperfect knowledge? More to the point, is there something that the agent should learn from the unexpected outcome that can be used in future decision making? In other words, why did the possible world $w(t)_i^*$ occur?

The answer to a why-question about an event in the possible world $w(t)_i^*$ requires an explanation. Inference to the best explanation has been called *abductive reasoning* to distinguish it from simple inductive generalization. Abductive reasoning can be defined with reference to a decision problem as the search for the most likely decision world $dw(t-1)_i^*$ from among the set of possible worlds $W(t-1)$, given the evidence of $w(t)_i^*$.

Decision analysis will therefore be useful to autonomous agents in contexts that use both forward predictive-based reasoning and backward explanation-based reasoning. When given a problem situation at time t, the agent should understand the causes and enabling conditions at t-1 in order to make decisions that will maximize utility at t+1. The next two sections formulate the elements of a domain independent account of causal structures that will be useful in organizing causal relations in the agent's knowledge base.

## 3. CAUSAL EXPLANATION IN A DETERMINISTIC UNIVERSE WITH PERFECT INFORMATION

The goal of the next two sections is to develop a theory of causal knowledge that can be used to constrain the search among possible worlds for an appropriate decision structure or explanation. These sections extend the causal analysis in [Star 1988a, 1988b] further. The theoretical structure should be useful for organizing existing causal knowledge as well as providing control knowledge about the most valuable kind of information to add to an incomplete database of causal relations. The causal theories presented here draw on the literature on explanation and causality in the philosophy of science. I have taken an eclectic approach, integrating aspects from several philosophers of science. The analysis in this section of a deterministic universe has been inspired mostly by Hume [1748/1955] and Mackie[1974].

*Contiguity, priority, and constant conjunction.* In artificial intelligence, there is a strong tradition of basing causal reasoning on a deterministic view of the universe. By deterministic I mean that it is assumed that every event is the result of causally sufficient conditions. This is a view of what the world is like, not a view about the knowledge we can obtain about the world. Even if the fundamental causal structure of the world is deterministic, references to probable cause will be appropriate if we are ignorant of some part of the causal complex--some "hidden factors" that, if understood, would complete the causal story.

I will, however, assume in this section that knowledge is based on perfect information: no noise, a complete causal theory, and no uncertainty. This is only adopted as a temporary convenience to allow us to proceed "as if" information is perfect. All the elements will be put in place to allow for generalizing the framework in the next section so that it can deal with difficult problems involving probable cause and uncertainty.

David Hume, the British empiricist philosopher, has been one of the most influential early writers on causality. He studied causality by considering the most obvious of situations, one billiard ball striking another. He says [1748/1955]:

> Here is a billiard ball lying on the table, and another ball moving toward it with rapidity. They strike; the ball which was formerly at rest now acquires a motion. This is as perfect an instance of the relation of cause and effect as any which we know either by sensation or reflection.

Hume found that three elements occur for all causes. The first element is *contiguity* or closeness in time and place between the cause and effect. It is often stated as "No action at a distance." Second, he found that causes have *priority* in time. No effect can occur before its cause. The third circumstance he called *constant conjunction*: Every object like the cause always produces some object like the effect. Hume

327

concluded, "Beyond these three circumstances of contiguity, priority, and constant conjunction I can discover nothing in this cause."

What Hume meant by this last statement was that the only way we seem to "know" that one billiard ball striking another will cause the second to move is because of past experience that it has always occurred thusly. In order to prove analytically that it will occur again it would be necessary to rely on the continuing regularity over time of the nature of causal events in our universe. But this regularity is only known from observation of past regularity; certainly we cannot assume what we are trying to prove. Thus, Hume concluded, the inference from cause to effect is based on a psychological expectation based on habit or custom rather than on some "real" connection.

Accepting Hume's argument means we must reject the view that the cause of some event can be determined by some objective, logically correct proof. If there is a logical proof linking cause and effect, it only shows that no inconsistency has occurred in the form of the argument. We cannot assign any causal meaning to the terms of the proof unless we are willing to do metareasoning about causality, which will require having a causal theory that relies on heuristic judgments of causal features.

*Context, causal scenarios, and the field*. Hume's analysis is only a first step. It leaves unanswered numerous problems. For example, suppose we want to explain why John woke up at seven in the morning. What was the cause of this event? Few people would say that the cause of John's waking up is his going to sleep the night before. But if John typically works nights and always wakes up to an alarm, then the event to be explained is why John woke up in the morning rather than the evening and not why his sleep was interrupted at seven o'clock. Clearly, an explanation of an event must be situated in a well-defined context. As van Fraassen [1980] has emphasized, the same event can be explained different ways if the context is different.

Suppose that a decision agent is faced with a specific decision problem that requires understanding why some undesirable event occurred and taking a decision to ensure that it does not recur. Let us focus on the elements of a causal explanation.

When most people judge that A has caused Z, they rarely mean that A is either necessary or sufficient to cause Z. Usually they mean that A together with a set of other specific conditions will be sufficient to cause Z. Define the conjunction of events sufficient to cause Z as a *causal scenario*. This brings us immediately to the problem of trying to determine which variables belong in a causal scenario. It also requires a distinction to be made within a scenario between causes and conditions.

Those events we call causes and conditions should be seen as differences that stand out against a background that helps define the problem context. Define the *field* to be the background of statements that are assumed to be true in a given context but that are omitted from the causal scenario. The single factor that distinguishes whether an event belongs in the field or in the scenario is whether or not the event should be considered as a possible cause in the given context. If it is not a possible cause but is relevant to the overall context, then it belongs in the field. If it is to be considered as a possible cause, then it belongs in the scenario. The *context* is therefore defined by the triple {*effect, scenario, field*}. Of course, if the context changes, then what was once an element of the field can become part of the scenario.

Let me introduce an example to illustrate these concepts. This example also forms the basis of a program that is mentioned later in the paper. Suppose a firm has a new cereal it plans to introduce in grocery stores. A new-product report is made that estimates the after-tax profits for eight periods after introducing the product. The report creates scenarios with optimistic and pessimistic assumptions about tax rates, discount rates, consumer spending, and store response. The report recommends introducing the cereal. After eight periods an evaluation is made and it is determined that the cereal is a flop. The president wants to know why.

Among the statements that are true and that might have entered into the report for this cereal are statements such as "cereal is usually eaten with cold milk for breakfast", "there are many competing brands", "shelf space is limited", "this cereal was made in a factory in Chicago", and that "tax rates are .4 times pretax profits". But these are statements that could be made about many of the cereals that the company has successfully introduced into the market. When the president is looking for an answer to his why-question, he wants to know why it is that this time the new product failed and other times, when cereals were introduced, they succeeded. Thus the question of failure or success is seen as a difference against a field of elements that do not enter as causal factors. More must be known before we can determine which specific factors belong in the field.

Suppose that the difference-in-a-background is that this time the cereal did not receive enough shelf space so that consumers did not notice it. We can elaborate the story by supposing that in the successful introductions the distributor supplied a special display stand that provided four feet of shelf space while this time only 10 inches was available, the width of one box. We thus see that statements such as "cereal is eaten with cold milk in the morning" or "there are many competing brands" can be set aside as belonging to the field. Given the context they are ruled out as candidates for the role of cause.

*Causes and inus conditions.* Among those facts that are part of the causal explanation, people usually distinguish between *proximate* or *triggering causes* and *standing conditions.* Suppose that in the cereal example the calculation of the net present value of after-tax profits from the new product depends on two other key conditions: the tax rate and the discount rate. If the tax rate had been higher, the product would not have been profitable even with the optimistic sale projections. If a lower discount rate had been used, even the realized low sales of cereal would have allowed the product to be a success.

The two conditions concerning the appropriate tax and discount rates are of a different nature than the condition requiring adequate shelf space. They should be classified as standing conditions that were true when the product failed. The tax rate is, in these circumstances, a parameter not under the control of the firm. The new-project discount rate is, in the given context, based on a policy that is independent of the particular new product. For example, the firm may require an after-tax rate of return on new projects of at least 20%. The context suggests that this is a standing condition when an executive makes a decision about launching a new product. The only variable factor is adequate shelf space. Thus the distinction between a cause and a condition also depends on the context. Since the agent is in a decision context and is searching for factors under its control that can be used to control the outcome, those events that are part of the causal scenario but not controllable are treated as conditions. In another context, say a scientific experiment, the scientist will try to control the conditions and let the "causal factor" vary. In both contexts, a single factor is nominated as the causal factor. If it does not provide the appropriate explanation, one looks at the conditions to see if in fact they varied. If so, the, *ex post*, a factor previously thought of as a condition should be treated as a cause.

We can go further in our causal analysis, following an approach first set out by Mackie [1974]. Let us continue to use the example of the introduction of a new cereal. Define the following variables: $A$ is insufficient shelf space, $B$ is the firm's discount rate, $C$ is the current tax rate, and $Z$ is the product's failed introduction. We have a scenario $ABC$ that produces an event $Z$. Let the causal field $F$ represent the statements describing the introduction of the new product and define the scenario $S$ as the conjunction of the three factors $ABC$ conditional on the field: $S = (ABC/F)$.

Define $S$ to be a *minimally sufficient scenario* since the conjunct of the three elements $A,B,C$ will always produce the effect, $Z$, the failed new product introduction. If a counter-factual situation is considered in which one of the conjuncts did not occur, the new cereal would have been a success. Each of these factors is, therefore, by itself an *insufficient* and *non-redundant* part of an *unnecessary* but *sufficient* scenario for $Z$. It will be convenient to say, using the first letters of the italicized words, that these factors are *inus* or satisfy the inus condition.

Now the statement that $A$ caused $Z$ means that

- $A$ is at least inus for $Z$;
- $A$ occurred;
- the other conjuncts $B$ and $C$ occurred; and
- all other minimally sufficient scenarios for $Z$ not having $A$ in them were absent on the occasion in question.

In other words, if $A$ is a cause of $Z$, $A$ is necessary for $Z$ to occur in the context described by the scenario and the field. We do not require that $A$ be sufficient in the context for $Z$ to occur. Moreover, if we cannot rule out definitively the existence of every other minimally sufficient scenario, the event $Z$ is overdetermined. In the situation of causal overdetermination due to incomplete knowledge any causal attribution must be a judgment of probable cause.

The elements of a causal theory in a deterministic universe with prefect information are now in place. A cause is related to its effect by contiguity, priority, and constant conjunction. The context of an event-to-be-explained is determined by the field and the minimally sufficient causal scenario. Irrelevant events are seen to be of two different types: those that have nothing to do with the problem and those that form part of the field. Events in the field are irrelevant in the sense that the should not be considered as conditions or possible causes. The scenario contains those events that are either standing conditions or triggering causes. The statement $A$ causes $Z$ has been defined in terms of inus conditions and scenarios.

The problem with this approach is that it has no room for limited knowledge, noise, statistical relations, and indeterminism. Nonetheless, it has provided us with a basic idealized model that is useful for understanding and organizing the causal relations in a domain independent way. We now turn to the more difficult problem of causal theory in an uncertain universe.

## 4. CAUSAL EXPLANATION IN AN UNCERTAIN UNIVERSE

In this section a theory of causal explanation will be presented that can be used in situations that involve indeterminism, partial knowledge, noisy events, and limited computing power. The theory follows closely the philosophical line of thought of Wesley Salmon [1984]. For other theories of causality that involve statistical or probabilistic notions see Reichenbach 1956; Carnap 1950; Greeno 1970; Hempel and Oppenheim 1948; Good 1983; van Fraassen 1980; and Suppes 1984]. Due to space limitations, only certain key points of the total theory will be discussed.

*The inverse relationship between prediction and explanation.* Perhaps the most widely accepted view of scientific explanation is that of Hempel and Oppenheim [1948]. They require an explanation to have predictive capability. They state, "...an explanation of a particular event is not fully adequate unless its explanans, if taken account of in time, could have served as a basis for predicting the event in question." They view low probability events as lacking this potential predictive force:

> Thus, we may be told that a car turned over on the road "because" one of its tires blew out while the car was traveling at high speed. Clearly, on the basis of just this information, the accident could not have been predicted, for the explanans provides no explicit general laws by means of which the prediction might be effected, nor does it state adequately the antecedent conditions which would be needed for the prediction. (p. 13)

Suppes [1984] also states a similar view, although he is not discussing explanation per se:

> ...there is a whole range of cases in which we do not have much hope of applying in an interesting scientific or commonsense way probabilistic analysis, because the causes will be surprising....Thus, although a Bayesian in such matters of individual events...I confess to being unable to make good probabilistic causal analyses of many kinds of individual events. (pp.64-65)

If we were to accept these views, an autonomous agent would be severely limited in its ability to make causal sense of low probability events. For example, if we draw one card at random from an ordinary 52-card deck, the probability of a spade or a club or a diamond is 3/4. The probability of the ace of spades is 1/52. Is it reasonable to say that the explanation of why we get a non-heart more adequate or better than the explanation of why we get the ace of spades?

The answer has been provided by Salmon [1984]:

> If determinism is false, a given set of circumstances of type C will yield an event of type E in a certain percentage of cases; and under circumstances of precisely the same type, no event of the type E will occur in a certain percentage of cases. If C defines a homogeneous reference class...then circumstances C explain the occurrence of E in those cases in which it occurs, and exactly the same circumstances C explain the nonoccurrence of E in those cases in which it fails to occur. The pattern is a statistical pattern, and precisely the same circumstances that produce E in some cases produce non-E in others. (p. 120)

Thus although low-probability events may be difficult to predict specifically, their occurrence a small percentage of the time is highly predictable. Thus the power of explanations is independent of the ability to predict individual events. This has important implications for an autonomous decision-making agent.

To illustrate the relevance for decision making, suppose that an autonomous agent is responsible for monitoring instruments on a Mars Rover vehicle (a wheeled vehicle for exploring the surface of Mars) and for making decisions when the vehicle encounters problems. Let us assume that while rolling around on martian soil, about once every 10,000 milliseconds its wheels will slip and spin freely for about 200 milliseconds, which will show up on an instrument in terms of a high rate of spin. Although the actual time index of the two milliseconds of slip cannot be predicted, the average number of milliseconds of slip per minute of rolling is a low probability event that is highly predictable. Statisticians would speak of this in terms of sufficient statistics. The specific time sequence of spin rates cannot be predicted but the average value is a sufficient statistic to describe the state of the machine. The agent need not look further than the sufficient statistic in building its explanation. Both the high-probability event of rolling without slippage and the low-probability event of a two-millisecond spin are considered normal and are "explained" equally well.

*Statistical relevance relation.* Recall that Hume searched unsuccessfully for the a true causal relationship, but only found contiguity, priority, and constant conjunction. The relation of constant conjunction was illustrated by one billiard ball hitting another and always causing the second to move. There was no uncertainty in the situation. Except for artificially simple problems and mathematical problems, there is almost always some degree of uncertainty involved in a complex cause-effect relation. For an autonomous agent, it should be assumed that the cause-effect relationship is not known with certainty. It is likely that the agent will have to deal with uncertain evidence in the form of a statistical relevance (S-R) relation, which is defined as follows:

*An event C is statistically relevant to the occurrence of event B in the context A if and only if*

(1) $P(B/AC)$ is not equal to $P(B/A)$; or  (2) $P(B/AC)$ is not equal to $P(B/A,\text{not-}C)$.

For a context A in which C occurs with a nonvanishing probability conditions (1) and (2) are equivalent.

The S-R relation makes explicit the need to refer to at least two probabilities for judging the relevance of two events: a prior probability and at least one or more posterior probabilities. But the S-R relation forms only a statistical basis for an explanation. It must be supplemented with causal factors in order to further constrain the space of possible explanations.

*Causal processes.* Suppose we have two events that are separated in space and time. Label event A the cause and event Z the effect. Let A be prior in time and statistically relevant to Z. Hume cited contiguity as a factor linking causal events. Yet in this case contiguity is missing. We need a *causal connection* to link together the two events in a causal relation. Salmon calls this link a *causal process*. He states:

> In some cases, such as the starting of the car, there are many intermediate events, but in such cases, the successive intermediate events are connected to one another by spatiotemporally continuous causal processes. [Salmon 1984, p. 156]

The reader is referred to Salmon for further details on causal processes.

*A common cause and conjunctive forks.* In influence diagrams and Bayes networks a distinction is made between causal factors and evidentiary factors through their spatial orientation [Shacter and Heckerman 1987; Pearl 1986, 1987]. It is also standard practice, when two factors X and Y are caused by a single factor C, to describe X and Y as being conditionally independent given the factor C. It is said that C *screens off* X from Y. We also say that C is the *common cause* of X and Y.

Hans Reichenbach [1956, sec. 19] recognized that a natural principle of causal reasoning is to expect events that are simultaneous, spatially separate, and strongly correlated, to have a common cause, prior in time, that generates the correlation. Reichenbach introduced the notion of a *conjunctive fork* in order to try to characterize the structure of relations involving common cause. It is defined in terms of the following four necessary conditions:

(1) $P(XY/C) = P(X/C)P(Y/C)$   (2) $P(XY/\text{not-}C) = P(X/\text{not-}C)P(Y/\text{not-}C)$

(3) $P(X/C) > P(X/\text{not-}C)$   (4) $P(Y/C) > P(Y/\text{not-}C)$.

These conditions also entail

(5) $P(XY) > P(X)P(Y)$   non-independence

(6) $P(X/C) = P(X/YC)$   conditional independence

(7) $P(Y/C) = P(Y/XC)$   conditional independence.

For a discussion see [Salmon 1984, pp. 160-161; Suppes 1984]. From condition (1) we see that given a common cause, X and Y are independent of each other. The same two factors, in the absence of their common cause C (and assuming no other common cause), are independent by virtue of (2). Yet the two factors are more highly correlated than they would be under strict independence; they are thus not independent, as shown in condition (5). Therefore, the occurrence of the event C makes X and Y statistically irrelevant to each other, and the occurrence of not-C has the same effect. Note that it is not claimed that any event C that fulfills relations (1)-(4) is a common cause of X and Y. C must be connected to X and Y by a causal process.

It is possible to describe the conjunctive fork in terms of a common effect E. Simply replace C by E in all the equations. This would correspond to some symptom, say a cough, being caused by two different illnesses, say a cold X or an allergy Y. Reichenbach claimed that there is an important asymmetry in conjunctive forks. He proposed that *it is impossible to have a situation in which two events, X and Y, in the absence of a common cause C, jointly produce a common effect E*. Let us call this Reichenbach's asymmetry conjecture or just *R-asymmetry*. Salmon has made a stronger claim. He proposes [1984, pp. 166-167] that *it is impossible to have a situation in which two events, X and Y, regardless of whether or not they have a common cause C, jointly produce a common effect E*. Refer to this as Salmon's asymmetry conjecture or just *S-asymmetry*.

Equations (1)-(7) plus the conditions of R-asymmetry and S-asymmetry provide important guides to the kind of probabilistic and statistical relations that an intelligent agent should look for in the data when trying to determine causal relations. They also suggest important restrictions on how causal relations can evolve over time and often furnish insights that are not immediately apparent. Space limitations prevent me from providing illustrations here.

## 5. TESTING THE THEORY

A theory of the structure of causal knowledge has been presented in order to suggest the kinds of relations that should be entered into an autonomous agent's database of knowledge and beliefs. The need for additional constraints comes about because the generation of the appropriate decision structure or explanation from completely unstructured data is an intractable problem. The value of the theory is, in this context, an empirical question. Does the addition of causal relations to a database enhance the performance of a decision-making agent? Which elements of the theory are most valuable, and what can be expected in terms of increased efficiency when they are used?

I have started to test the theories with some very simple programs. Some interesting results have occurred when techniques from machine learning have been used to generate explanations. For example, a program has been developed that generates a plan to introduce a new cereal on the market if prior analysis shows that the introduction will be profitable. A structure corresponding to an influence diagram is generated and the decision is made to introduce the cereal. Several periods after the introduction, additional data are introduced to the database. The data indicate that the cereal is not profitable. The program generates an explanation of why the cereal is not profitable, doing goal regression to find the sufficient preconditions for profitability.

The program generates structures corresponding to two different influence diagrams. The first is a decision structure that terminates in a successful product introduction. The second is an explanation structure that shows the preconditions that led to the unsuccessful product introduction. By comparing the two structures, the program should be able to determine the differences and thus find the cause of the failure. The problem is that at the most detailed level, since we are dealing with numeric data, the values at almost every node are different in the two structures. The solution is to abstract from the values by generalizing until only qualitative or binary values remain. By tracing back and comparing nodes we can arrive at the conclusion that a single different ground proposition is responsible for the different outcomes. In order for this to succeed, the values that correspond to standing conditions in a causal scenario have to be protected. For example, the tax rate and the discount rate are taken as standing conditions that do not change value.   If facts are identified as irrelevant, as belonging to the field, or as conditions, the number of unifications needed for the program to arrive at an answer is reduced substantially. Since the example is still quite simple, the program always finds a solution. But the number of operations needed to get to the solution drops continually as more causal knowledge is introduced into the database.

## 6. CONCLUSIONS AND FUTURE RESEARCH

This paper has attempted to bring together some aspects of the symbolic approaches of artificial intelligence and machine learning with the more quantitative approach of decision analysis. It is a continuation of previous work that I have done towards integrating the two approaches [Star 1987]. The problem of generating the appropriate decision structure and causal explanations relies heavily on the symbolic computing tradition of artificial intelligence. But the use of decision analysis, probabilities, and measurable utility is characteristic of quantitative methods. Perhaps one of the most important conclusions that I would like readers to reach after reading this paper is that symbolic and quantitative approaches are necessary complements to each other--both contribute to the tentative answers I have proposed about how to create a structured representation of a decision problem faced by an autonomous decision-making agent.

The theory of explanation and of the causal structure of the world has only been sketched out quite briefly. It is not meant to provide a definitive answer to the questions that philosophers ask about scientific explanation and causality; rather it draws on the philosophical literature for ideas that can be useful for solving computational problems. Thus the theory needs to be tested extensively by using it in examples and in working programs to find both its weaknesses and its strengths. Our very preliminary results are encouraging, but much additional work needs to be done.

## 7. ACKNOWLEDGEMENTS

I would like to thank Neeraj Bhatnagar for comments on section 2 and an anonymous referee for comments on the first draft of this paper. All remaining errors are my responsibility.